\documentclass[11pt,a4paper]{article}

\usepackage[final]{acl}
\usepackage{times}
\usepackage{latexsym}

\usepackage{amsmath}
\usepackage{amsfonts}
\usepackage[linesnumbered,ruled,vlined]{algorithm2e}
\usepackage{graphicx} % DO NOT CHANGE THIS
\usepackage{booktabs}
\usepackage{array}
\usepackage{enumitem}
\usepackage{amsthm}
\usepackage{algpseudocode}
\usepackage[switch]{lineno}
\usepackage[utf8]{inputenc}
\usepackage{eqparbox}
\usepackage[nopar]{lipsum}
\usepackage{multirow}
\usepackage{makecell}
\usepackage{xcolor}
\usepackage{hhline} 

\usepackage{microtype}

\usepackage{relsize}

\usepackage{booktabs,multirow,array}
\newcolumntype{N}{@{}m{0pt}@{}}%a fix for array package

\usepackage[many]{tcolorbox}
\newtcolorbox{fancyquotes}{%
    enhanced jigsaw, 
    breakable,      % allow page breaks
    frame hidden,   % hide the default frame
    left=0.5cm,       % left margin
    right=0.1cm,      % right margin
    overlay={%
        \node [scale=8,
            text=black,
            inner sep=0pt,] at ([xshift=-1cm,yshift=-1cm]frame.north west){}; 
        \node [scale=8,
            text=black,
            inner sep=0pt,] at ([xshift=1cm]frame.south east){};  
            },
                parbox=false,
}

\usepackage{mathtools, nccmath}
\usepackage{scrextend}
\deffootnote[.25in]{.25in}{.15in}{\makebox[.25in][r]{\thefootnotemark .\hspace{.15in}}}

\makeatletter

\newtheorem*{proof*}{Proof}

\usepackage{listings}
\usepackage{color}
\definecolor{codegreen}{rgb}{0.3,0.5,0.0}
\def\@fnsymbol#1{\ensuremath{\ifcase#1\or \dagger\or *\or \ddagger\or
   \mathsection\or \mathparagraph\or \|\or **\or \dagger\dagger
   \or \ddagger\ddagger \else\@ctrerr\fi}}

\newcolumntype{C}[1]{>{\centering\let\newline\\\arraybackslash\hspace{0pt}}m{#1}}
\newcommand\ChangeRT[1]{\noalign{\hrule height #1}}

\title{RetroMAE: Pre-Training Retrieval-oriented Language Models Via Masked Auto-Encoder}

\author{Shitao Xiao$^1$\thanks{The two researchers make equal contributions to this work and are designated as co-first authors.} , Zheng Liu$^2$\footnotemark[1], Yingxia Shao$^1$, Zhao Cao$^2$\\
  1: Beijing University of Posts and Telecommunications, Beijing, China \\ 
  2: Huawei Technologies Ltd. Co., Shenzhen, China \\
  \texttt{\{stxiao,shaoyx\}@bupt.edu.cn}, 
  \texttt{\{liuzheng107,caozhao1\}@huawei.com}
}

\begin{document}
\maketitle

\begin{abstract}
Despite pre-training's progress in many important NLP tasks, it remains to explore effective pre-training strategies for dense retrieval. In this paper, we propose \textbf{RetroMAE}, a new retrieval oriented pre-training paradigm based on Masked Auto-Encoder (MAE). RetroMAE is highlighted by three critical designs. 1) \textbf{A novel MAE workflow}, where the input sentence is polluted for encoder and decoder with different masks. The sentence embedding is generated from the encoder's masked input; then, the original sentence is recovered based on the sentence embedding and the decoder's masked input via masked language modeling. 2) \textbf{Asymmetric model structure}, with a full-scale BERT like transformer as encoder, and a one-layer transformer as decoder. 3) \textbf{Asymmetric masking ratios}, with a moderate ratio for encoder: 15$\sim$30\%, and an aggressive ratio for decoder: 50$\sim$70\%. Our framework is simple to realize and empirically competitive: {the pre-trained models dramatically improve the SOTA performances on a wide range of dense retrieval benchmarks, like \textbf{BEIR} and \textbf{MS MARCO}. The source code and pre-trained models are made publicly available at https://github.com/staoxiao/RetroMAE so as to inspire more interesting research.}

% after pre-trained on English Wikipedia and BookCorpus, RetroMAE achieves superior performances on BEIR ({zero-shot}), MS MARCO and Natural Question ({supervised}). 
% \blue{And adopting in-domain pretraining, RetroMAE achieves the state-of-the-art performance on MSMARCO dataset.}
% \blue{Codes and models are available at https://github.com/staoxiao/RetroMAE.}

% supervised and zero-shot performances on MS MARCO, Natural Question, and BEIR. 

% , where the input sentence is polluted on both encoder and decoder side with different masks, and original sentence is reconstructed based on both sentence embedding and masked sentence;
% 2) asymmetric model architectures, with a large-scale expressive transformer for sentence encoding and a extremely simplified transformer for sentence reconstruction; 3) asymmetric masking ratios, with a moderate masking on the encoder side (15\%) and an aggressivev masking ratio on the decoder side (50$\sim$90\%). We pre-train a BERT like encoder on English Wikipedia and BookCorpus, where it notably outperforms the existing pre-trained models on a wide range of dense retrieval benchmarks, like MS MARCO, Open-domain Question Answering, and BEIR. 
\end{abstract} 

%% pretrained language model
%% pretraining for dense retrieval
%% masked auto-encoder
%% what's the difference
%% anything tricky
%% experiment 

%% dense retrieval -> data dependency -> pretrained model -> retrieval oriented -> worth exploration -> motivation: strengthen the semantic encoding capability -> -> existing work: examples, generative & contrastive learning -> surprising well with neither complicated designs -> inspired by recent success on cv pretraining, a simple mae framework -> critical designs -> expressive encoder -> moderate masking -> embedding -> aggressive mask -> simplified decoder, reconstruct (no seq2seq) -> various empirical studies -> ms marco, odqa, beir 

\section{Introduction}
Dense retrieval is important to many web applications. By letting semantically correlated query and document represented as spatially close embeddings, dense retrieval can be efficiently conducted via approximate nearest neighbour search, such as PQ \cite{jegou2010product,xiao2021matching,xiao2022distill} and HNSW \cite{malkov2018efficient}. 
Recently, large-scale language models have been widely used as the encoding networks for dense retrieval \cite{karpukhin2020dense,xiong2020approximate,luan2021sparse}. The mainstream models, e.g., BERT \cite{Devlin2019BERT}, RoBERTa \cite{Liu2019Roberta}, T5 \cite{raffel2019exploring}, are usually pre-trained by token-level tasks, like MLM and Seq2Seq. However, the sentence-level representation capability is not fully developed in these tasks, which restricts their potential for dense retrieval. 

Given the above defect, there have been increasing interests to develop retrieval oriented pre-trained models. One popular strategy is to leverage self-contrastive learning \cite{chang2020pre,guu2020realm}, where the model is trained to discriminate positive samples from data augmentation. However, the self-contrastive learning can be severely limited by the data augmentation's quality; besides, it usually calls for massive amounts of negative samples \cite{he2020momentum,chen2020simple}. Another strategy relies on auto-encoding \cite{gao2021condenser,lu2021less,wang2021tsdae}, which is free from the restrictions on data augmentation and negative sampling. The current works are differentiated in how the encoding-decoding workflow is designed, and it remains an open problem to explore more effective auto-encoding framework for retrieval oriented pre-training. 

% which will severely influence their performances on downstream retrieval tasks. So far, it is still an open problem of exploring more effective AE based pre-training algorithms. 

\begin{figure}[t]
\centering
\includegraphics[width=0.99\linewidth]{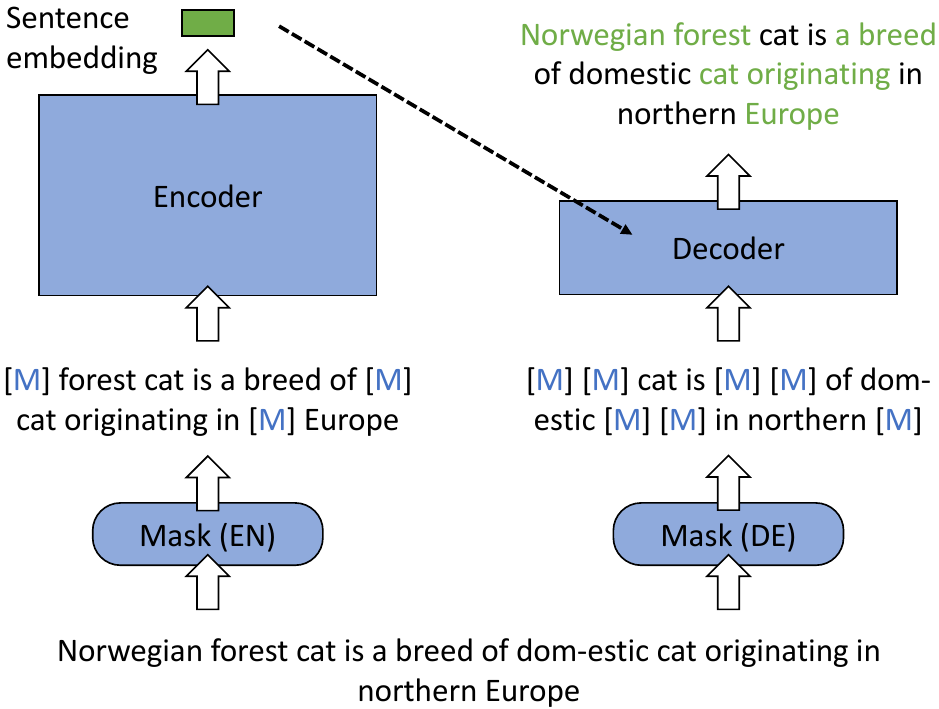}
\vspace{-15pt}
\caption{RetroMAE. The encoder utilizes a full-scale BERT, whose input is moderately masked. The decoder is a one-layer transformer, whose input is aggressively masked. The original input is recovered based on the sentence embedding and the decoder's input via MLM.} 
\vspace{-10pt}
\label{fig:1}
\end{figure}

We argue that two factors are critical for the auto-encoding based pre-training: 1) the reconstruction task must be demanding enough on encoding quality, 2) the pre-training data needs to be fully utilized. We propose RetroMAE (Figure \ref{fig:1}), which optimizes both aspects with the following designs. 

% the reconstruction difficulty and data efficiency are critical factors for the pre-training effectiveness of auto-encoding. Thus, we propose a novel framework called RetroMAE (Figure \ref{fig:1}), which optimizes the above two factors on top of the following designs. 

% The proposed framework is simple to realize, and the generated model is surprisingly competitive when fine-tuned for dense retrieval. 

% In this paper, we propose a novel masked auto-encoding (MAE) framework to pre-train retrieval-oriented language models, known as RetroMAE (Figure \ref{fig:1}). The proposed pre-training framework not only simplifies the existing AE based methods, but also gives rise to surprisingly competitive performances on downstream dense retrieval tasks. In particular, RetroMAE is featured for the following critical components and strategies. 

$\bullet$ \textbf{A novel MAE workflow}. The pre-training follows a novel masked auto-encoding workflow. The input sentence is polluted twice with two different masks. One masked input is used by encoder, where the sentence embedding is generated. The other one is used by decoder: joined with the sentence embedding, the original sentence is recovered via masked language modeling (MLM). 

$\bullet$ \textbf{Asymmetric structure}. RetroMAE adopts an asymmetric model structure. The encoder is a full-scale BERT, which is able to generate discriminative embedding for the input sentence. In contrast, the decoder follows an extremely simplified structure, i.e., a single-layer transformer, which is learned to reconstruct the input sentence. 

$\bullet$ \textbf{Asymmetric masking ratios}. The encoder's input is masked at a moderate ratio: 15$\sim$30\%, which is slightly above its traditional value in MLM. However, the decoder's input is masked at a much more aggressive ratio: 50$\sim$70\%. 

The above designs of RetroMAE are favorable to the pre-training effectiveness thanks to the following reasons. Firstly, the auto-encoding is made {\textbf{more demanding on encoding quality}}. The conventional auto-regression may attend to the prefix during the decoding process; and the conventional MLM only masks a small portion (15\%) of the input tokens. By comparison, RetroMAE aggressively masks most of the input for decoding.
% \textit{To ensure the reconstruction quality, the model is forced to encode in-depth semantics within its sentence embedding}. 
% Besides, the decoder is merely a one-layer transformer; the extremely simplified network further increases the reconstruction difficulty.
% \blue{
% To ensure the reconstruction quality,
% the decoder is forced to rely on the knowledge in the sentence embedding, and the encoder is forced to encode in-depth semantics within its sentence embedding.}
{As such, the reconstruction will be not enough to leverage the decoder's input alone, but heavily depend on the sentence embedding. Thus, it will force the encoder to capture in-depth semantics of the input}. 
Secondly, it ensures \textbf{training signals to be fully generated} from the input sentence. For conventional MLM-style methods, the training signals may only be generated from 15\% of the input tokens. Whereas for RetroMAE, the training signals can be derived from the majority of the input. Besides, knowing that the decoder only contains one-single layer, we further propose the \textbf{enhanced decoding} on top of two-stream attention \cite{yang2019xlnet} and position-specific attention mask \cite{dong2019unified}. As such, 100\% of the tokens can be used for reconstruction, and each token may sample a unique context for its reconstruction.  

% forcing the in-depth semantics to be encoded within the sentence embedding so as to ensure the reconstruction quality. Besides, the decoder is merely a one-layer transformer; the extremely simplified network further increases the difficulty of auto-encoding. Secondly, it ensures \textbf{training signals to be fully generated} from each pre-training sentence. For typical MLM style methods, the training signals may only be derived from 15\% of the input tokens. Whereas for RetroMAE, the training signals can be derived from the majority of the tokens. Besides, knowing that the decoder only contains one-single layer, we propose the \textbf{enhanced decoding} with two-stream attention \cite{yang2019xlnet} and position-specific attention mask \cite{dong2019unified}, where the training signals can be derived from the entire input tokens. 

{RetroMAE is simple to realize and empirically competitive. We merely use a moderate-amount of data (Wikipedia, BookCorpus, MS MARCO corpus) for pre-training, where a BERT base scale encoder is learned. For the zero-shot setting, it produces an average score of \textbf{45.2 on BEIR} \cite{thakur2021beir}; and for the supervised setting, it may easily reach an MRR@10 of \textbf{41.6 on MS MARCO} passage retrieval \cite{nguyen2016ms} following standard knowledge distillation procedures. Both values are unprecedented for dense retrievers with the same model size and pre-training conditions. We also carefully evaluate the impact introduced from each of the components, whose results may bring interesting insights to the future research.}

% We use English Wikipedia and BookCorpus for pre-training, from which a BERT base scale encoder is produced. We evaluate our pre-trained model comprehensively: with its zero-shot performance evaluated on BEIR benchmark \cite{thakur2021beir}, and  the supervised performance evaluated with MS MARCO \cite{nguyen2016ms} and Natural Question \cite{kwiatkowski2019natural}. 
% the supervised performances evaluated on MS MARCO \cite{nguyen2016ms} and Natural Question \cite{kwiatkowski2019natural}, and its zero-shot performances evaluated on BEIR \cite{thakur2021beir}. 
% According to our experiments, RetroMAE outperforms both generic and retrieval oriented pre-trained models with notable advantages. 

% We perform comprehensive experimental studies with popular dense retrieval benchmarks, such as MS MARCO \cite{nguyen2016ms} and BEIR \cite{thakur2021beir}. According to the evaluation results, RetroMAE notably improves both in-domain and out-of-domain performance in comparison with the existing generic and retrieval-oriented pretrained language models. 

\section{Related works}
% The related works are reviewed from two aspects: dense retrieval and pretrained language models.

Dense retrieval is widely applied to web applications, like search engines \cite{karpukhin2020dense}, advertising \cite{lu2020twinbert,zhang2022uni} and recommender systems \cite{xiao2022training}. It encodes query and document within the same latent space, where relevant documents to the query can be efficiently retrieved via ANN search. The encoding model is critical for the retrieval quality. Thanks to the recent development of large-scale language models, e.g., BERT \cite{Devlin2019BERT}, RoBERTa \cite{Liu2019Roberta}, and T5 \cite{raffel2019exploring}, there has been a major leap-forward for dense retrieval's performance \cite{karpukhin2020dense,luan2021sparse,lin2021pretrained}. 

The large-scale language models are highly differentiated in terms of pre-training tasks. One common task is the masked language modeling (MLM), as adopted by BERT \cite{Devlin2019BERT} and RoBERTa \cite{Liu2019Roberta}, in which the masked tokens are predicted based on their context. The basic MLM is extended in many ways. For example, tasks like entity masking, phrase masking and span masking \cite{sun2019ernie,joshi2020spanbert} may help the pre-trained models to better support the sequence labeling applications, such as entity resolution and question answering. Besides, tasks like auto-regression \cite{radford2018improving,yang2019xlnet} and Seq2Seq \cite{raffel2019exploring,lewis2019bart} are also utilized, where the pre-trained models are enabled to serve NLG related scenarios. However, most of the generic pre-trained models are based on token-level tasks, where the sentence representation capability is not effectively developed \cite{chang2020pre}. Thus, it may call for a great deal of labeled data \cite{nguyen2016ms,kwiatkowski2019natural} and sophisticated fine-tuning methods \cite{xiong2020approximate,qu2020rocketqa} to ensure the pre-trained models' performance for dense retrieval. 

To mitigate the above problem, recent works propose retrieval oriented pre-trained models. The existing methods can be divided as the ones based on self-contrastive learning (SCL) and the ones based on auto-encoding (AE). The SCL based methods \cite{chang2020pre,guu2020realm,xu2022laprador} rely on data augmentation, e.g., inverse cloze task (ICT), where positive samples are generated for each anchor sentence. Then, the language model is learned to discriminate the positive samples from the negative ones via contrastive learning. However, the self-contrastive learning usually calls for huge amounts of negative samples, which is computationally expensive. Besides, the pre-training effect can be severely restricted by the quality of data augmentation. The AE based methods are free from these restrictions, where the language models are learned to reconstruct the input sentence based on the sentence embedding. The existing methods utilize various reconstruction tasks, such as MLM \cite{gao2021condenser} and auto-regression \cite{lu2021less,wang2021tsdae,li2020optimus}, 
{which are highly differentiated in terms of how the original sentence is recovered and how the training loss is formulated. For example, the auto-regression relies on the sentence embedding and prefix for reconstruction; while MLM utilizes the sentence embedding and masked context. The auto-regression derives its training loss from the entire input tokens; however, the conventional MLM only learns from the masked positions, which accounts for 15\% of the input tokens. Ideally, we expect the decoding operation to be demanding enough, as it will force the encoder to fully capture the semantics about the input so as to ensure the reconstruction quality. Besides, we also look forward to high data efficiency, which means the input data can be fully utilized for the pre-training task.} 

% \blue{which are highly differentiated in terms of data efficiency and reconstruction difficulty. For example, the conventional MLM only learns from the masked positions, which accounts for 15\% of the input tokens; in contrast, the auto-regression may learn from the entire tokens, which leads to better data efficiency. Besides, the conventional auto-encoding leverages equally sized encoder and decoder \cite{li2020optimus,wang2021tsdae}; while in \cite{lu2021less}, a smaller decoder is used, which increases the reconstruction difficulty. Ideally, the auto-encoding should be data efficient, such that the pre-training corpus can be fully utilized; meanwhile, the reconstruction task should be made sufficiently challenging, which forces the encoder to capture in-depth semantics about the input sentences. }

% the training signals can be derived from all the input tokens with auto-regression. Besides, the conventional auto-encoding leverages a large-scale decoder \cite{li2020optimus,wang2021tsdae}; while in \cite{lu2021less}, a smaller decoder is used, which increases the reconstruction difficulty. Ideally, the auto-encoding should be data efficient, which ensures the pre-training corpus to be fully leveraged; meanwhile, it should also be made sufficiently challenging, which forces sentence embeddings to capture the in-depth semantics about the input sentences.  

%% basic workflow: auto-encoding: masked input sentence for encoder -> sentence embedding -> masked input for decoder -> decoding output
%% enhancement with position-specific attention mask 

\section{Methodology}
We develop a novel masked auto-encoder for retrieval oriented pre-training. The model contains two modules: a BERT-like encoder $\Phi_{enc}(\cdot)$ to generate sentence embedding, and a one-layer transformer based decoder $\Phi_{dec}(\cdot)$ for sentence reconstruction. The original sentence $X$ is masked as $\tilde{X}_{enc}$ and encoded as the sentence embedding $\mathbf{h}_{\tilde{X}}$. The sentence is masked again (with a different mask) as $\tilde{X}_{dec}$; together with $\mathbf{h}_{\tilde{X}}$, the original sentence $X$ is reconstructed. Detailed elaborations about RetroMAE are made as follows. 

\subsection{Encoding}\label{sec:encoding}
\begin{figure*}[t]
\centering
\includegraphics[width=1.0\textwidth]{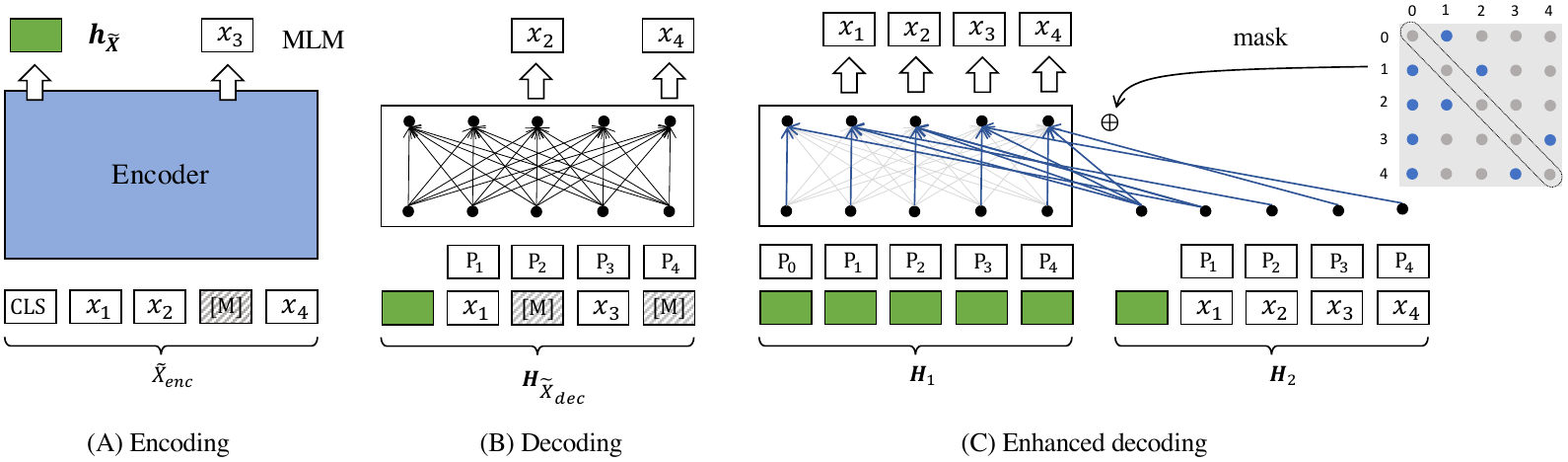}
% \vspace{-15pt}
\caption{RetroMAE pre-training workflow. (A) Encoding: the input is moderately masked and encoded as the sentence embedding (the green rectangle). (B) Decoding: the input is aggressively masked, and joined with the sentence embedding to reconstruct the masked tokens (the shadowed tokens). (C) Enhanced encoding: all input tokens are reconstructed based on the sentence embedding and the visible context in each row (defined in Eq. \ref{eq:7}); the main diagonal positions are filled with $-\infty$ (grey), and positions for the visible context are filled with $0$ (blue).}  
% \vspace{-7pt}
\label{fig:2}
\end{figure*}

The input sentence $X$ is polluted as $\tilde{X}_{enc}$ for the encoding stage, where a small fraction of its tokens are randomly replaced by the special token [M] (Figure \ref{fig:2}. A). We apply a moderate masking ratio (15$\sim$30\%), which means the majority of information about the input will be preserved. Then, the encoder $\Phi^{enc}(\cdot)$ is used to transform the polluted input as the sentence embedding $\mathbf{h}_{\tilde{X}}$: 
\begin{equation}
    \mathbf{h}_{\tilde{X}} \leftarrow \Phi_{enc}(\tilde{X}_{enc}). 
\end{equation}
We apply a BERT like encoder with 12 layers and 768 hidden-dimensions, which helps to capture the in-depth semantics of the sentence. Following the common practice, we select the [CLS] token's final hidden state as the sentence embedding. 

% \blue{
% \begin{equation}
% \begin{gathered}
% \hat{\mathbf{H}}_{\tilde{X}_{dec}} =
% [\mathbf{h}_{\tilde{X}};\mathbf{H}_{\tilde{X}_{dec}}[1:N]].
% \end{gathered}
% \end{equation}}
% \blue{where $\mathbf{H}_{\tilde{X}_{dec}}$ is the input representation (summary of token embedding and position embedding) for sentence $\tilde{X}_{dec}$. And we replace the embedding of first token (i.e., [CLS]) with the sentence embedding from encoder.}
% $x_i$ equals to the original token value if it is not masked, otherwise [M].

% \begin{equation}
%     % \max. P(X|\mathbf{h}_{\tilde{X}} \oplus \mathbf{E}_{\tilde{X}_{dec}} + \mathbf{p}).
%     \min. \sum_{x_i \in \text{masked}} \mathrm{CE}(x_i|\mathbf{h}_{\tilde{X}} \oplus \mathbf{E}_{\tilde{X}_{dec}} + \mathbf{P}),
% \end{equation} 

\subsection{Decoding}\label{sec:decode}
The input sentence $X$ is polluted as $\tilde{X}_{dec}$ for the decoding stage (Figure \ref{fig:2}. B). The masking ratio is more aggressive than the one used by the encoder, where 50$\sim$70\% of the input tokens will be masked. The masked input is joined with the sentence embedding, based on which the original sentence is reconstructed by the decoder. Particularly, the sentence embedding and the masked input are combined into the following sequence: 
\begin{equation}
    \mathbf{H}_{\tilde{X}_{dec}} \leftarrow [\mathbf{h}_{\tilde{X}}, \mathbf{e}_{x_1}+\mathbf{p}_1, ... , \mathbf{e}_{x_N}+\mathbf{p}_N]. 
\end{equation}
In the above equation, $\mathbf{e}_{x_i}$ denotes the embedding of $x_i$, to which an extra position embedding $\mathbf{p}_i$ is added. 
Finally, the decoder $\Phi_{dec}$ is learned to reconstruct the original sentence $X$ by optimizing the following objective: 
\begin{equation}
    \mathcal{L}_{dec} = \sum_{x_i \in \text{masked}} \mathrm{CE}(x_i|\Phi_{dec}( \mathbf{H}_{\tilde{X}_{dec}}) ),
\end{equation}
where $\mathrm{CE}$ is the cross-entropy loss. As mentioned, we use a one-layer transformer based decoder. Given the aggressively masked input and the extremely simplified network, the decoding becomes challenging, which forces the generation of high-quality sentence embedding so that the original input can be recovered with good fidelity. 

% $\mathbf{h}_{\tilde{X}} \oplus \mathbf{E}_{\tilde{X}_{dec}} + \mathbf{P}$
\subsection{Enhanced Decoding}\label{sec:enhance}
One limitation about the decoding process is that the training signals, i.e., the cross-entropy loss, can only be derived from the masked tokens. Besides, every masked token is always reconstructed based on the same context, i.e., $\mathbf{H}_{\tilde{X}_{dec}}$. We argue that the pre-training effect can be further enhanced providing that 1) \textbf{more training signals} can be derived from the input sentence, and 2) the reconstruction task can be performed based on \textbf{diversified contexts}. To this end, we propose the \textbf{enhanced decoding} inspired by two-stream self-attention \cite{yang2019xlnet} and position-specific attention mask \cite{dong2019unified}. Particularly, we generate two input streams: $\mathbf{H}_1$ (query) and $\mathbf{H}_2$ (context), for the decoding operation (Figure \ref{fig:2}. C):
\begin{equation}
\begin{gathered}
\label{eq:4}
\mathbf{H}_1 \leftarrow [\mathbf{h}_{\tilde{X}} + \mathbf{p}_0,...,
\mathbf{h}_{\tilde{X}} + \mathbf{p}_N], \\
\mathbf{H}_2 \leftarrow  
[\mathbf{h}_{\tilde{X}}, \mathbf{e}_{x_1}+\mathbf{p}_1, ... , \mathbf{e}_{x_N}+\mathbf{p}_N]. 
% \mathbf{H}_2 \leftarrow  
% [\mathbf{h}_{\tilde{X}};\mathbf{H}_{X}[1:N]],
\end{gathered}
\end{equation}
where $\mathbf{h}_{\tilde{X}}$ is the sentence embedding, $\mathbf{e}_{x_i}$ is the token embedding (no token is masked in this place), and $\mathbf{p}_i$ is the position embedding.
We introduce the position-specific attention mask $\mathbf{M} \in \mathbb{R}^{L \times L}$, where the self-attention is performed as:
\begin{equation}\label{eq:5}
\begin{gathered}
    \mathbf{Q} = \mathbf{H}_1\mathbf{W}^Q, \mathbf{K} = \mathbf{H}_2\mathbf{W}^K, \mathbf{V} = \mathbf{H}_2\mathbf{W}^V; \\
    \mathbf{M}_{ij} = 
    \begin{cases}
    0, ~~~~~~\text{can be attended}, \\
    -\infty, ~\text{masked}; 
    \end{cases} \\
    \mathbf{A} = \mathrm{softmax}(\frac{\mathbf{Q}^T\mathbf{K}}{\sqrt{d}} + \mathbf{M
    })\mathbf{V}. 
\end{gathered}
\end{equation}
The output $\mathbf{A}$, together with $\mathbf{H}_{1}$ (because of the residual connection) are used to reconstruct the original input (other operations, like layer-norm and FFN, are omitted from our discussion). Finally, the following objective will be optimized: 
% \begin{equation}\label{eq:6}
%     \min. \sum_{x_i \in X} \mathrm{CE}(x_i|\mathbf{A}, \mathbf{H}_{1}, \mathbf{M}).
% \end{equation} 
\begin{equation}\label{eq:6}
 \mathcal{L}_{dec} = \sum_{x_i \in X} \mathrm{CE}(x_i|\mathbf{A}, \mathbf{H}_{1}).
\end{equation} 
Knowing that the decoder only consists of one single transformer layer, each token $x_i$ is reconstructed based on the context which are visible to the $i$-th row of matrix $\mathbf{M}$. In this place, the following rules are applied to generate the position specific attention mask matrix $\mathbf{M}$: 
\begin{equation}\label{eq:7}
\begin{gathered}
\mathbf{M}_{ij} = 
    \begin{cases}
    0,   ~~ x_j \in s(X_{\neq i}), ~\text{or}~ j_{{|i\neq0}}=0 \\
    -\infty,  ~~ \text{otherwise}. 
    \end{cases} 
\end{gathered}
\end{equation}
% In the above equation, $s(X_{\neq i})$ represents the random sampling of the input tokens. 
The sampled tokens, $s(X_{\neq i})$, and the 1st position (except for the 1st row) will be visible when reconstructing $x_i$. The diagonal elements, i.e., $x_i$ for the $i$-th row, will always be excluded, which means they will always be masked; as a result, each token cannot attend to itself during the reconstruction. 

% In particular, the main diagonal elements ($j=i$) will always be $-\infty$, such that each token may not attend to itself during the reconstruction; the first column is excluded from masking ($X_{\neq0}$) such that the sentence embedding will always be attended; the other positions are randomly masked (denoted by the $s(\cdot)$ operator), where the total amount of masked tokens in each row may reach the pre-defined masking ratio. 

We summarize the pre-training workflow with the enhanced decoding as Algorithm \ref{alg:1}. Note that the original masked language modeling task in BERT is kept in encoder. The corresponding loss, denoted as $\mathcal{L}_{enc}$, is added with the decoder's loss, which formulates the final training loss.
The following features need to be emphasized for our pre-training workflow. Firstly, the reconstruction task is \textbf{demanding} given the aggressive masking ratio and the extremely simplified network of decoder. Secondly, we may derive \textbf{abundant} pre-training signals from the unsupervised corpus since all tokens within each input sentence can be used for the reconstruction. Finally, the pre-training is \textbf{simple} to realize: 1) there are no requirements on sophisticated data augmentation and negative sampling, and 2) the computation cost is maintained to be similar with the conventional BERT/RoBERTa style pre-training given the simplicity of decoder. 

\begin{algorithm}[t]
\caption{RetroMAE}\label{alg:1}
    \LinesNumbered 
    % \SetKwInOut{KwIn}{Input}
    % \SetKwInOut{KwOut}{Output}
    % \KwIn{...}
    % \KwOut{...} 
    \Begin{
        $\tilde{X}_{enc} \leftarrow mask(X)$\;
        $\mathbf{h}_{\tilde{X}} \leftarrow \Phi_{enc}(\tilde{X}_{enc})$\;
        % \blue{compute $\mathcal{L}_{enc}$} \;
        % $\mathbf{H}_1 \leftarrow \mathbf{h}_{\tilde{X}} + \mathbf{P}$;  ~~~ $\%$ for $\mathbf{Q}$\\
        % $\mathbf{H}_2 \leftarrow \mathbf{E}_{X} + \mathbf{P}$; ~~~ $\%$ for $\mathbf{K}$ and $\mathbf{V}$\\ 
        {$\mathbf{H}_1, \mathbf{H}_2 \leftarrow$ Eq. \ref{eq:4}\;}
        $\mathbf{M} \leftarrow$ Eq. \ref{eq:7}\;
        $\mathbf{A} \leftarrow$ based on $\mathbf{H}_1,\mathbf{H}_2,\mathbf{M}$ as Eq. \ref{eq:5}\;
        % model update w.r.t. Eq. \ref{eq:6}\; 
        {$\mathcal{L}_{dec} \leftarrow$ Eq.~\ref{eq:6}} \;
        {model update w.r.t. $\mathcal{L}_{enc}+\mathcal{L}_{dec} $}\;
    }
\end{algorithm}

\section{Experimental Studies} 
{We evaluate the retrieval performance of the sentence embedding generated by RetroMAE's pre-trained encoder, where two major issues are explored.} 1) RetroMAE's impact on zero-shot and supervised dense retrieval, in comparison with the generic pre-trained models and the retrieval oriented pre-trained models. 2) The impact from the four technical factors in RetroMAE: the enhanced decoding, the decoder size, the decoder's masking ratio, and the encoder's masking ratio. 
% \blue{}

\subsection{Experiment Settings}
% \subsubsection{Datasets.}
The following \textbf{datasets} are utilized for the pre-training and evaluation of RetroMAE.  

$\bullet$ \textbf{Pre-training}. We reuse the same pre-training corpora as the ones utilized by BERT \cite{Devlin2019BERT}: \textbf{\textbf{English Wikipedia}} and \textbf{BookCorpus}. Corresponding datasets are also frequently used by the previous works on retrieval oriented pre-training, such as SEED \cite{lu2021less} and Condenser \cite{gao2021condenser}. {Following coCondenser \cite{gao2021unsupervised}, we also use \textbf{MS MARCO} corpus to analyze the effect of in-domain pre-training for dense retrieval (which we find critical to the performance on MS MARCO but unnecessary to the performances on other datasets).}

$\bullet$ \textbf{Evaluation}. 
% \blue{In the downstream tasks, we discard the decoder and only use the encoder to generate sentence embedding.}
We use two datasets to evaluate the retrieval performance after supervision. 1) \textbf{MS MARCO} \cite{nguyen2016ms}, which contains queries from Bing Search. We use its passage retrieval task, which contains 502,939 training queries and 6,980 evaluation queries (Dev). The answer needs to be retrieved from 8.8 million candidate passages. 2) \textbf{Natural Questions} \cite{kwiatkowski2019natural}, which consists of queries from Google. There are 79,168 training queries, 8,757 dev queries, and 3,610 testing queries. The answer is retrieved from 21,015,324 Wikipedia passages. 
We evaluates the zero-shot retrieval performance on \textbf{BEIR benchmark} \cite{thakur2021beir}. It fine-tunes the pre-trained model with MS MARCO queries, and evaluate the zero-shot transferability on the other 18 datasets. The evaluation data covers dense retrieval tasks across different domains, such as question answering, fact checking, bio-medical retrieval, news retrieval, and social media retrieval, etc.

\begin{table*}[t]
    \centering
    % \small
    \scriptsize
    % \footnotesize
    \begin{tabular}{p{1.7cm}|C{0.9cm}|C{1.1cm}|C{1.1cm}|C{1.2cm}|C{1.0cm}|C{1.0cm}|C{0.9cm}|C{1.2cm}|C{1.2cm} }
    \ChangeRT{1pt}
    \textbf{Datasets} &
    \textbf{BERT} & \textbf{RoBERTa} & \textbf{DeBERTa} & \textbf{LaPraDoR} & \textbf{SimCSE} & \textbf{DiffCSE} & \textbf{SEED} & \textbf{Condenser} & \textbf{RetroMAE} \\
    \hline
    TREC-COVID & 0.615 & 0.649 & 0.688 & 0.478 & 0.460 & 0.492 & 0.627 & 0.750 & \textbf{0.772} \\
    BioASQ & 0.253 & 0.279 & 0.290 & 0.252 & 0.263 & 0.258 & 0.308 & 0.322 & \textbf{0.421} \\
    NFCorpus & 0.260 & 0.243 & 0.238 & 0.310 & 0.260 & 0.259 & 0.278 & 0.277 & \textbf{0.308}  \\
    \hline
    NQ & 0.467 & 0.413 & 0.452 & 0.454 & 0.435 & 0.412 & 0.446 & 0.486 & \textbf{0.518} \\
    HotpotQA & 0.488 & 0.448 & 0.474 & 0.513 & 0.502 & 0.499 & 0.541 & 0.538 & \textbf{0.635} \\
    FiQA-2018 & 0.252 & 0.291 & 0.299 & 0.288 & 0.250 & 0.229 & 0.259 & 0.259 & \textbf{0.316} \\
    \hline
    Signal-1M(RT) & 0.204 & 0.229 & 0.243 & 0.241 & 0.262 & 0.260 & 0.256 & 0.261 & \textbf{0.265} \\
    \hline
    TREC-NEWS & 0.362 & 0.385 & 0.378 & 0.286 & 0.356 & 0.363 & 0.358 & 0.376 & \textbf{0.428} \\
    Robust04 & 0.351 & 0.384 & 0.378 & 0.299 & 0.330 & 0.343 & 0.365 & 0.349 & \textbf{0.447} \\
    \hline
    ArguAna & 0.265 & 0.395 & 0.297 & \textbf{0.499} & 0.413 & 0.468 & 0.389 & 0.298 & 0.433 \\
    Touche-2020 & 0.259 & \textbf{0.299} & 0.271 & 0.137 & 0.159 & 0.168 & 0.225 & 0.248 & 0.237 \\
    \hline
    CQADupStack & 0.282 & 0.278 & 0.279 & 0.309 & 0.290 & 0.305 & 0.290 & \textbf{0.347} & 0.317 \\
    Quora & 0.787 & 0.509 & 0.846 & 0.837 & 0.844 & 0.850 & \textbf{0.852} & 0.853 & 0.847 \\
    \hline
    DBPedia & 0.314 & 0.275 & 0.271 & 0.334 & 0.314 & 0.303 & 0.330 & 0.339 & \textbf{0.390} \\
    \hline
    SCIDOCS & 0.113 & 0.111 & 0.106 & 0.150 & 0.124 & 0.125 & 0.124 & 0.133 & \textbf{0.150} \\
    \hline
    FEVER & 0.682 & 0.683 & 0.594 & 0.511 & 0.623 & 0.641 & 0.641 & 0.691 & \textbf{0.774} \\
    Climate-FEVER & 0.187 & 0.222 & 0.160 & 0.173 & 0.211 & 0.200 & 0.176 & 0.211 & \textbf{0.232} \\
    SciFact & 0.533 & 0.539 & 0.543 & 0.531 & 0.554 & 0.523 & 0.575 & 0.593 & \textbf{0.653} \\
    \hhline{=|=|=|=|=|=|=|=|=|=}
    AVERAGE & 0.371 & 0.368 & 0.378 & 0.367 & 0.369 & 0.372 & 0.391 & 0.407 & \textbf{0.452} \\
    \ChangeRT{1pt}
    \end{tabular}
    % \vspace{-5pt}
    \caption{Zero-shot dense retrieval performances on BEIR benchmark (measured by NDCG@10).} 
    % \vspace{-10pt}
    \label{tab:3}
\end{table*}

We consider three types of \textbf{baseline methods} in our experimental studies\footnote{For all baseline methods, we use their officially released pre-trained models for our experiments.}. 

$\bullet$ \textbf{Generic models}. The following generic pre-trained language models are used in our experiments. 1) \textbf{BERT} \cite{Devlin2019BERT}, which is the most popular pre-trained language model in practice. It is also frequently fine-tuned as the encoding network for dense retrievers \cite{karpukhin2020dense,xiong2020approximate}. 2) \textbf{RoBERTa} \cite{Liu2019Roberta}, which is an enhanced replication of BERT with substantially enlarged training data and improved training settings. 3) \textbf{DeBERTa} \cite{he2020deberta}, which further improves BERT and RoBERTa with disentangled attention mechanism and an enhanced mask decoder; it is one of the strongest pre-trained models on NLU benchmarks, such as GLUE and SuperGLUE. 

% introduces the generator-discriminator framework and the token replacement prediction task to further improve the pre-training effect. 

$\bullet$ \textbf{Retrieval oriented models}. We consider two types of retrieval oriented pre-trained models in our experiments. One is based on self-contrastive learning, where the following methods are included. 1) \textbf{SimCSE} \cite{gao2021simcse}, in which the language model is learned to discriminate different views of the anchor sentence augmented by drop-out. 2) \textbf{LaPraDoR} \cite{xu2022laprador}, an enhancement of conventional ICT pre-training \cite{guu2020realm,chang2020pre}; it proposes to train the query and document encoder iteratively so that the scale of negative samples can be expanded. 3) \textbf{DiffCSE} \cite{chuang2022diffcse}, which enhances SimCSE with the jointly utilization of self-contrastive learning and conditional difference prediction. The other one is based on auto-encoding, in which the following methods are included. 4) \textbf{Condenser} \cite{gao2021condenser}, where the sentence embedding is joined with the intermediate hidden-states from encoder for MLM. 5) \textbf{SEED} \cite{lu2021less}, where the sentence embedding is used to recover the original input via 
auto-regression.

$\bullet$ \textbf{Implementation details}. RetroMAE utilizes bi-directional transformers as its encoder, with 12 layers, 768 hidden-dim, and a 30522-token vocabulary (same as BERT base). The decoder is a one-layer transformer. The default masking ratios are 0.3 for encoder and 0.5 for decoder. The model is trained for 8 epochs, with AdamW optimizer, batch-size 32 (per device), learning rate 1e-4. The training is on a machine with 8$\times$ Nvidia A100 (40GB) GPUs. 
The models are implemented with PyTorch 1.8 and HuggingFace transformers 4.16. 
We adopt the official script \footnote{\url{https://github.com/beir-cellar/beir/blob/main/examples/retrieval/training/train\_msmarco\_v3.py}} from BEIR to prepare the models for their zero-shot evaluation. 
% We use \textbf{DPR} \cite{karpukhin2020dense} and \textbf{ANCE} \cite{xiong2020approximate} to fine-tune the pre-trained models for supervised evaluations. 
{For supervised evaluations, we use \textbf{DPR} \cite{karpukhin2020dense} and \textbf{ANCE} \cite{xiong2020approximate} to fine-tune the pre-trained models. We also evaluate the models' performance on MS MARCO with standard knowledge distillation. Particularly, we train one BERT base scale cross-encoder over hard negatives returned by the ANCE-finetuned bi-encoder; then, we further finetune the bi-encoder by minimizing the KL-divergence with the cross-encoder.}

% Following coCondenser\cite{gao2021unsupervised} and AR2~\cite{zhang2021adversarial}, we also conduct the experiments of RetroMAE-style pre-training on MSMARCO dataset and distilling the knowledge from cross-encoder.}

% Our models and source code will be released to public after review. 
% The well-trained models and source code will be released to public after review.  

\begin{table*}[t]
    \centering
    % \small
    \scriptsize
    % \footnotesize
    \begin{tabular}{p{1.2cm}|C{1.1cm}|C{1.2cm}|C{1.0cm}|C{1.0cm}|C{1.0cm}|C{0.9cm}|C{0.9cm}|C{0.9cm}|C{0.9cm}|C{1.0cm} }
    \ChangeRT{1pt} 
    & 
    \multicolumn{5}{c|}{\textbf{MS MARCO}} & \multicolumn{5}{c}{\textbf{Natural Question}} \\
    \cmidrule(lr){1-1}
    \cmidrule(lr){2-6}
    \cmidrule(lr){7-11}
    \textbf{Methods} & 
    \textbf{MRR@10} & \textbf{MRR@100} & \textbf{R@10} & \textbf{R@100} & \textbf{R@1000} & \textbf{R@10} & \textbf{R@20} & \textbf{R@30} & \textbf{R@50} & \textbf{R@100} \\
    \hline
    BERT & 0.3170 & 0.3278 & 0.5801 & 0.8570 & 0.9598 & 0.7399 & 0.7925 & 0.8136 & 0.8396 & 0.8668 \\
    RoBERTa & 0.3136 & 0.3258 & 0.5638 & 0.8478 & 0.9579 & 0.7150 & 0.7676 & 0.7939 & 0.8211 & 0.8476 \\
    DeBERTa &0.3186 & 0.3304 & 0.5824 & 0.8625 & 0.9654 & 0.7152 & 0.7778 & 0.8022 & 0.8269 & 0.8510 \\
    \hline
    SimCSE & 0.3193 & 0.3307 & 0.5907 & 0.8653 & 0.9699 & 0.7291 & 0.7864 & 0.8125 & 0.8391 & 0.8670 \\ 
    LaPraDoR & 0.3191 & 0.3307 & 0.5833 & 0.8537 & 0.9602 & 0.7377 & 0.7920 & 0.8155 & 0.8399 & 0.8677 \\
    % \hline
    DiffCSE & 0.3202 & 0.3311 & 0.5832 & 0.8561 & 0.9607 & 0.7393 & 0.7934 & 0.8155 & 0.8407 & 0.8673 \\
    SEED & 0.3264 & 0.3374 & 0.5913 & 0.8535 & 0.9539 & 0.7454 & 0.7958 & 0.8208 & 0.8432 & 0.8701 \\
    Condenser & 0.3357 & 0.3471 & 0.6082 & 0.8770 & 0.9683 & 0.7562 & 0.8053 & 0.8269 & 0.8501 & 0.8711 \\
    \hhline{=|=|=|=|=|=|=|=|=|=|=}
    RetroMAE & \textbf{0.3553} & \textbf{0.3665} & \textbf{0.6356} & \textbf{0.8922} & \textbf{0.9763} & \textbf{0.7704} & \textbf{0.8172} & \textbf{0.8399} & \textbf{0.8604} & \textbf{0.8812} \\
    \ChangeRT{1pt}
    \end{tabular}
    % \vspace{-5pt}
    \caption{Supervised evaluation results based on DPR fine-tuning.} 
    % \vspace{-5pt}
    \label{tab:1}
\end{table*}

\begin{table*}[t]
    \centering
    % \small
    \scriptsize
    % \footnotesize
    \begin{tabular}{p{1.2cm}|C{1.1cm}|C{1.2cm}|C{1.0cm}|C{1.0cm}|C{1.0cm}|C{0.9cm}|C{0.9cm}|C{0.9cm}|C{0.9cm}|C{1.0cm} }
    \ChangeRT{1pt} 
    & 
    \multicolumn{5}{c|}{\textbf{MS MARCO}} & \multicolumn{5}{c}{\textbf{Natural Question}} \\
    \cmidrule(lr){1-1}
    \cmidrule(lr){2-6}
    \cmidrule(lr){7-11}
    \textbf{Methods} & 
    \textbf{MRR@10} & \textbf{MRR@100} & \textbf{R@10} & \textbf{R@100} & \textbf{R@1000} & \textbf{R@10} & \textbf{R@20} & \textbf{R@30} & \textbf{R@50} & \textbf{R@100} \\
    \hline
    BERT & 0.3460 & 0.3569 & 0.6220 & 0.8734 & 0.9642 & 0.7875 & 0.8227 & 0.8460 & 0.8601 & 0.8776 \\
    RoBERTa & 0.3433 & 0.3543 & 0.6130 & 0.8705 & 0.9637 & 0.7629 & 0.8053 & 0.8277 & 0.8449 & 0.8698\\
    DeBERTa & 0.3396 & 0.3512 & 0.6016 & 0.8719 & 0.9670 & 0.7654 & 0.8097 & 0.8288 & 0.8479 & 0.8698 \\
    \hline
    SimCSE & 0.3520 & 0.3623 & 0.6276 & 0.8849 & 0.9738 & 0.7742 & 0.8194 & 0.8418 & 0.8626 & 0.8864 \\ 
    LaPraDoR & 0.3456 & 0.3564 & 0.6129 & 0.8755 & 0.9640 & 0.7801 & 0.8247 & 0.8424 & 0.8590 & 0.8773 \\
    % \hline
    DiffCSE & 0.3462 & 0.3571 & 0.6217 & 0.8748 & 0.9654 & 0.7853 & 0.8252 & 0.8410 & 0.8620 & 0.8784 \\
    SEED & 0.3544 & 0.3653 & 0.6263 & 0.8812 & 0.9687 & 0.7803 & 0.8258 & 0.8449 & 0.8684 & 0.8870 \\
    Condenser & 0.3635 & 0.3742 & 0.6388 & 0.8912 & 0.9722 & 0.7903 & 0.8325 & 0.8524 & 0.8668 & 0.8834 \\
    \hhline{=|=|=|=|=|=|=|=|=|=|=} 
    RetroMAE & \textbf{0.3822} & \textbf{0.3928} & \textbf{0.6677} & \textbf{0.9074} & \textbf{0.9807} & \textbf{0.8044} & \textbf{0.8443} & \textbf{0.8632} & \textbf{0.8776} & \textbf{0.8942} \\
    \ChangeRT{1pt}
    \end{tabular}
    % \vspace{-5pt}
    \caption{Supervised evaluation results based on ANCE fine-tuning.} 
    % \vspace{-5pt}
    \label{tab:2}
\end{table*}

\begin{table}[t]
    \centering
    % \small
    \scriptsize
    % \footnotesize
    \begin{tabular}{|p{1.6cm}|C{1.1cm}|C{0.9cm}|C{0.9cm}|C{1.0cm}| }
    \ChangeRT{1pt} 
    & 
    \multicolumn{4}{c|}{\textbf{MS MARCO}}  \\
    \hline
    \textbf{Methods} & 
    \textbf{MRR@10} &  \textbf{R@10} & \textbf{R@100} & \textbf{R@1000}  \\
    \hline
    coCondenser & 0.382 & - & - & 0.984  \\
    \hline
    RetroMAE & \textbf{0.393} & \textbf{0.675} & \textbf{0.918} & \textbf{0.985}   \\
    \ChangeRT{1pt}
    \end{tabular}
    % \vspace{-5pt}
    \caption{RetroMAE vs. coCondenser on MS MARCO. Both models are fine-tuned by ANCE.} 
    % \vspace{-5pt}
    \label{tab:domain}
\end{table}  

\begin{table}[t]
    \centering
    % \small
    \scriptsize
    % \footnotesize
    \begin{tabular}{|p{1.6cm}|C{1.1cm}|C{0.9cm}|C{0.9cm}|C{1.0cm}| }
    \ChangeRT{1pt} 
    & 
    \multicolumn{4}{c|}{\textbf{MS MARCO}}  \\
    \hline
    \textbf{Methods} & 
    \textbf{MRR@10} &  \textbf{R@10} & \textbf{R@100} & \textbf{R@1000}  \\
    \hline
    AR2 & {0.395} & {-} & {-} & {0.986}   \\
    ColBERTv2 & {0.397} & {-} & {-} & {0.984}   \\
    RocketQAv2 & {0.388} & {-} & {-} & {0.981}   \\
    ERNIE-Search & {0.401} & {-} & {-} & {0.982}   \\
    \hline
    RetroMAE & \textbf{0.416} & \textbf{0.709} & \textbf{0.927} & \textbf{0.988}   \\
    \ChangeRT{1pt}
    \end{tabular}
    % \vspace{-5pt}
    \caption{RetroMAE vs. the recent dense retrievers; all models are fine-tuned by knowledge distillation.} 
    \vspace{-5pt}
    \label{tab:distill}
\end{table} 

\subsection{Main Results} 
We analyze the zero-shot performances in Table \ref{tab:3}, where RetroMAE achieves remarkable advantages: it produces the best empirical results on most of the datasets, and it surpasses the strongest baseline by \textbf{+4.5\%} in total average. The leap-forward of zero-shot performance is much larger than any of the improvements made by the baseline models. The improvement is indeed thrilling knowing that it is not from the increasing of model scale or the enrichment of pre-training data, but purely from the upgrade of pre-training algorithm. 
We further demonstrate the supervised evaluations in Table \ref{tab:1} and \ref{tab:2}, where the pre-trained models are fine-tuned with DPR and ANCE. The baselines are partitioned into two groups according to whether they are generic pre-trained models or retrieval oriented ones. It can be observed that RetroMAE maintains notable advantages over the baselines. With DPR fine-tuning, it outperforms the strongest baselines by +1.96\% (MRR@10) and +1.48\% (Recall@10) on the two datasets; with ANCE fine-tuning, corresponding advantages become +1.42\% (MRR@10) and +1.41\% (Recall@10). 

{As for RetroMAE's \textbf{pre-trained models over MS MARCO} corpus. With ANCE fine-tuning (Table~\ref{tab:domain}), our method outperforms its peer approach coCondenser \cite{gao2021unsupervised} by \textbf{+1.1\%} on MRR@10 (which is of the same model size and same pre-training data). While under the knowledge distillation scenario (Table~\ref{tab:distill}), RetroMAE surpasses a series of strong baselines proposed in the recent period, including the models with highly sophisticated distillation methods: AR2 \cite{zhang2021adversarial} by \textbf{+2.1\%}, RocketQAv2 \cite{ren2021rocketqav2} by \textbf{+2.8\%}, ERNIE-search \cite{lu2022ernie} by \textbf{+1.5\%}, and the late-interaction model ColBERTv2 \cite{santhanam2021colbertv2} by \textbf{+1.9\%}.} 

% \blue{Besides, we further pre-train the RetroMAE on MSMARCO dataset and fintune it by ANCE (Table~\ref{tab:domain}) and distillation (Table~\ref{tab:distill}). We can observe that RetroMAE outperforms both coCondenser and AR2, achieving the state-of-the-art performance in MSMARCO dataset.And compared with the results in Table~\ref{tab:2}, RetroMAE has a significantly improvement, which shows the effictiveness of in-doman RetroMAE-style pre-training.}

To summarize, the above results verify RetroMAE's effectiveness from two aspects. 1) It substantially improves the pre-trained model's \textbf{transferability}, which helps to result in superior zero-shot performances on out-of-domain datasets. 2) It provides a \textbf{strong initialization of dense retriever}; after fine-tuned with in-domain data, it gives rise to a high-quality supervised retrieval performance in the corresponding scenario. Besides the primary results, we may also have the following interesting observations.

\begin{table*}[t]
    \centering
    % \small
    \scriptsize
    % \footnotesize
    \begin{tabular}{C{1.5cm}|C{1.5cm}|C{1.1cm}|C{1.2cm}|C{1.0cm}|C{1.0cm}|C{1.1cm}|C{1.2cm}|C{1.0cm}|C{1.0cm} }
    \ChangeRT{1pt}
    & &
    \multicolumn{4}{c|}{\textbf{MS MARCO}} & \multicolumn{4}{c}{\textbf{Natural Questions}} \\
    \cmidrule(lr){1-1}
    \cmidrule(lr){2-2}
    \cmidrule(lr){3-6}
    \cmidrule(lr){7-10}
    \textbf{Factor} & \textbf{Setting} & 
    \textbf{MRR@10} & \textbf{MRR@100} & \textbf{R@100} & \textbf{R@1000} & 
    \textbf{MRR@10} & \textbf{MRR@100} & \textbf{R@100} & \textbf{R@1000} \\
    \hline
    \multirow{2}{*}{Decoding}
    & w. enhance & \textbf{0.3553} & \textbf{0.6356} & \textbf{0.8922} & \textbf{0.9763} & \textbf{0.7704} & \textbf{0.8399} & \textbf{0.8604} & \textbf{0.8812} \\
    & w.o. enhance & 0.3462 & 0.6218 & 0.8813 & 0.9725 & 0.7562 & 0.8291 & 0.8540 & 0.8759 \\
    \hline
    \multirow{3}{*}{\begin{tabular}{c}Decoder\\
                                      size (w.o.)
                \end{tabular}} 
    & $\#\text{layer}=1$ & {0.3462} & 0.6218 & 0.8813 & 0.9725 & 0.7562 & {0.8291} & {0.8540} & 0.8759 \\
    & $\#\text{layer}=2$ & 0.3446 & 0.6217 & 0.8828 & 0.9729 & 0.7561 & 0.8289 & 0.8538 & 0.8759 \\ 
    & $\#\text{layer}=3$ & 0.3439 & {0.6223} & {0.8829} & {0.9730} & {0.7563} & 0.8290 & 0.8537 & {0.8760} \\
    \hline
    \multirow{6}{*}{\begin{tabular}{c}Mask ratio \\
                                       (decoder)
                \end{tabular}}
    & $0.15$ (w.) & 0.3496 & 0.6297 & 0.8905 & 0.9734 & 0.7608 & 0.8309 & 0.8554 & 0.8750 \\
    & $0.5$ (w.)  & \textbf{0.3553} & \textbf{0.6356} & \textbf{0.8922} & \textbf{0.9763} & \textbf{0.7704} & \textbf{0.8399} & \textbf{0.8604} & \textbf{0.8812} \\
    & $0.9$ (w.)  & 0.3514 & 0.6285 & 0.8905 & 0.9740 & 0.7609 & 0.8343 & 0.8562 & 0.8756 \\
    \cmidrule(lr){2-10}
    & $0.15$ (w.o.) & 0.3440 & 0.6177 & 0.8802 & 0.9700 & 0.7519 & 0.8253 & 0.8523 & 0.8758 \\
    & $0.7$ (w.o.) & \textbf{0.3508} & \textbf{0.6262} & \textbf{0.8850} & \textbf{0.9738} & \textbf{0.7593} & \textbf{0.8327} & \textbf{0.8551} & \textbf{0.8760} \\
    & $0.9$ (w.o.) & 0.3441 & 0.6198 & 0.8803 & 0.9725 & 0.7576 & 0.8307 & \textbf{0.8551} & 0.8745 \\
    \hline 
    \multirow{3}{*}{\begin{tabular}{c}Mask ratio \\
                                       (encoder)
                    \end{tabular}}
    & $0.15$ (w.) & 0.3501 & 0.6306 & 0.8890 & 0.9757 & 0.7703 & \textbf{0.8404} & \textbf{0.8604} & 0.8795 \\
    & $0.3$ (w.) & \textbf{0.3553} & \textbf{0.6356} & \textbf{0.8922} & \textbf{0.9763} & \textbf{0.7704} & 0.8399 & \textbf{0.8604} & \textbf{0.8812} \\
    & $0.9$ (w.) & 0.3365 & 0.6143 & 0.8750 & 0.9701 & 0.7599 & 0.8296 & 0.8508 & 0.8692 \\ 
    \ChangeRT{1pt}
    \end{tabular}
    % \vspace{-5pt}
    \caption{Ablation studies. (``w.''/``w.o.'' indicates ``with''/``without'' using the enhanced decoding.)} 
    \vspace{-6pt}
    \label{tab:4}
\end{table*}

Firstly, the advanced pre-training methods in generic areas do not necessarily contribute to the dense retrieval performances. Particularly, both RoBERTa and DeBERTa are major improvements of BERT; however, none of them is able to achieve better performances than BERT as they did on general NLU benchmarks. This observation further supports the motivation to develop retrieval oriented pre-trained models. 

Secondly, the auto-encoding style pre-training (adopted by SEED, Condenser, and RetroMAE) is empirically proved to be much more favorable for dense retrieval, given its dominance over the generic and self-contrastive learning based pre-trained models in both zero-shot evaluation and supervised evaluation.  

Thirdly, the self-contrastive learning based pre-trained models bring very little improvements over the generic ones when fine-tuning is made available, as their performances are close to each other in both supervised and zero-shot evaluations.
% \footnote{{LaPraDoR's performance is lowered than its originally claimed result; however, after confirmed with its author, our reproduction (based on its officially released checkpoint) reflects its real performances in our settings.}}. 
% \footnote{Although LaPraDoR claimed better results, they were found to be falsely reported after confirmed with its author. Our reproduction reflects its true performance.}. 
In fact, there is no supervise about this observation considering the similar results reported by recent studies on dense retrieval \cite{gao2021condenser} (BERT against ICT) and image processing \cite{el2021large} (BEiT \cite{bao2021beit} against MoCo/DINO). 
That is the pre-trained models from self-contrastive learning tend to have comparatively limited potential for fine-tuning. Therefore, we argue that the retrieval-oriented pre-training algorithm should be properly designed according to the condition of fine-tuning in specific scenarios. 

% should always consider whether quality fine-tuning data is available. 

% where models based on self-contrastive learning were observed to be limited for fine-tuning. 

% With the observations from our work and the related studies, we argue that self-contrastive learning and other algorithms, especially the ones based on auto-encoding, should be properly considered for retrieval-oriented pre-training, according to whether quality fine-tuning data is available. 

% Secondly, the contrastive learning based methods merely bring very limited improvements over the generic pre-trained models, as can be observed from the comparison between SimCSE, LaPraDoR and BERT in Table \ref{tab:1} and \ref{tab:2}. In fact, similar observations are also made by previous study \cite{gao2021condenser}: although contrastive learning may equip the pre-trained models with preliminary capability on dense retrieval, the advantage is almost wiped out when the models are fine-tuned with labeled data.  

% Thirdly, despite that RoBERTa and ELECTRA are proved to be more effective than BERT on generic NLU tasks, like GLUE and MRC, they are no better than BERT on dense retrieval scenarios. Such an observation validates once again that the conventional token-level pre-training contributes little to the models' dense retrieval capability; thus, retrieval-oriented pre-training is needed.  

% although contrastive learning may equip the pre-trained models with preliminary capability on dense retrieval, the advantage is almost wiped out when the models are fine-tuned with labeled data. 

\subsection{Ablation Studies}
%% enhanced > basic, all position used for decoding -> more training signals -> better training outcome
%% smaller decoder better than lager decoder
%% larger decoder ratio better than smaller decoder ratio
%% larger encoder ratio also benefits the performance but smaller than decoder ratio
%% higher reconstruction difficulty, force model to generate in-depth sentence embedding from encoder 

We ablate RetroMAE with its supervised performance (DPR fine-tuning) in Table \ref{tab:4}, where the following factors are analyzed: 1) decoding method, 2) decoder's size, 3) decoder's masking ratio, 4) encoder's masking ratio. We may have the following observations from our results. 

Firstly of all, we analyze the impact from the \textbf{decoding method}, i.e., whether the enhanced decoding is used or not. It can be observed that the enhanced decoding (w.) notably outperforms the basic decoding (w.o.). Such an empirical advantage can be attributed to the improved data efficiency of the enhanced decoding. Under the default masking ratio, the basic decoding merely samples 50\% of the input tokens for reconstruction, all of which are predicted based on the same context. With the enhanced decoding (Section \ref{sec:enhance}), all of the input tokens can be utilized for reconstruction, and each of the tokens is reconstructed based on a unique context sampled as Eq. \ref{eq:7}. As such, the enhanced decoding may obtain more sufficient and diversified training signals from the input data. 

Secondly, we analyze the impact from \textbf{decoder size}, with the number of transformer layers increased from 1 to 3. Knowing that the enhanced decoding is only applicable for single-layer transformers, it is disabled for these experiments. Despite higher computation costs, the enlarged decoders do not bring any empirical gains. Besides, considering that the enhanced decoding has to be disabled for large decoders (\#layer$>$1), which will severely harm the retrieval performances, the one-layer decoder is proved to be the best option. 

% . In this place, we use two different decoders for comparison: 1) the decoder with one-single transformer layer ($H_{de}=1$), and 2) the decoder with two transformer layers ($H_{de}=2$). It can be found that the smaller decoder, which increases the difficulty of input reconstruction, gives rise to better empirical performances. 

Thirdly, we make an analysis for different \textbf{masking ratios of decoder}, whose value is increased from 0.15 to 0.9. We introduce two sets of experiments: one with the enhanced decoding (w.), and the other one without using enhanced decoding (w.o.). For both experiments, we observe substantial improvements of retrieval quality resulted from the aggressive masking ratios. For enhanced decoding (w.), the optimal performance is achieved at 0.5; without using enhanced decoding (w.o.), the optimal performance is reached at 0.7. This minor difference is probably because: unlike ``w.'' where all input tokens can be reconstructed, ``w.o.'' only reconstructs the masked tokens; thus, it may trade a larger ratio for the increasing of training signals. 

Lastly, we study the \textbf{encoder's masking ratio}. It is quite interesting that a slightly improved masking ratio of 0.3 also improves the empirical performances, compared with the commonly used value 0.15. For both encoder and decoder, the increased reconstruction difficulty can be the common reason why the increased masking ratios benefit the retrieval quality. However, different from decoder, a too aggressive ratio of encoder, e.g., 0.9, will severely harm the retrieval performance. This is because a too large masking ratio will prevent the generation of high-quality sentence embedding, considering that most of the useful information of the input sentence will be discarded. 

% achieves better performances than the default one $\gamma_{en}=0.15$. Similar as the decoder's situation, an increased masking ratio on the encoder side will also increase the reconstruction difficulty. However, the empirical performance will not benefit from an even larger masking ratio; and the ideal value of $\gamma_{en}$ is smaller than $\gamma_{de}$. This is because a too large $\gamma_{en}$ will prevent the generation of high-quality sentence embedding, considering that too much useful information about the input sentence will be discarded. 

The ablation studies are \textbf{concluded as follows}: 1) RetroMAE's performance can be notably improved by the enhanced decoding; 2) the one-layer transformer is the best for decoder; 3) the retrieval quality can be improved from an aggressive masking ratio of decoder, and a moderately improved masking ratio of encoder. 

% based on our experimental results. Firstly, RetroMAE's effectiveness is verified by its notable advantages on supervised and zero-shot evaluations. Secondly, the 

% Firstly, the auto-encoding framework demonstrates strong potential in pre-training retrieval-oriented language models, and RetroMAE brings in substantially improvements over the existing auto-encoding based methods. Secondly, RetroMAE's performance is optimized by the enhanced decoding strategy, the simplified network of decoder, and the proper setting of the masking ratios.  

% It can be found that the enhanced decoding outperforms the basic decoding with notably advantages. Such an observation indicates that the pre-training effect can be substantially benefit from the improved data efficiency, i.e., the 

\section{Conclusion}
We propose RetroMAE, a novel masked auto-encoding framework to pre-train retrieval oriented language models: the input sentence is randomly masked for encoder and decoder, and the sentence embedding is joined with the decoder's masked input to reconstruct the original input. We introduce an asymmetric model structure (full-scale encoder and single-layer decoder) and asymmetric masking ratios (a moderate ratio for encoder and an aggressive one for decoder), which makes the reconstruction sufficiently demanding. We also introduce the enhanced decoding, which makes the full utilization of the pre-training data. Our experiments on BEIR, MS MARCO, and Natural Question validate RetroMAE's effectiveness, as significant improvements on both zero-shot and supervised evaluations can be achieved over the existing methods.

% The proposed framework is highlighted for the following properties: 1) asymmetric structure: a full-scale encoding network and a single-layer transformer based decoding network; 2) asymmetric masking ratios: a moderate masking ratio for encoder and an aggressive masking ratio for decoder. We further introduce the enhanced decoding based on 

% \clearpage 
\section{Limitations} {So far, our empirical studies are performed based on BERT base scale transformers; meanwhile, merely a moderate amount of pre-training data is used (mainly due to the limitations on computation resources). Despite the demonstrated effectiveness, it's still necessary to explore the impact from enlarged networks and increased pre-training data, as both factors were found to be important in recent works \cite{ni2021large}.} 

% Besides, given that our models are pre-trained on large-scale corpus, it might encounter the similar risk of bias and toxic patterns in generic language models.

% Besides, our method could be a generic framework to pre-train sentence embeddings; its potential for more scenarios beyond dense retrieval, e.g., SentEval, SentGLUE, remains to be explored in the future.
%  same biases and toxic

\section*{Acknowledgements}
This work is supported by the National Natural Science Foundation of China (Nos. U1936104, 62272054, 62192784), and CCF-Tencent Open Fund.

% In this work, we propose the RetroMAE and conduct extensive experiments on both finetune and zero-shot scenarios. However, these
% results are reported based on the base-sized models, lacking of more results for different sizes of model. The performance of larger-scale model is not verified and we will investigate it in the future. Besides, more sentence-level tasks, such as sentence similarity, also need to be studied to explore potential application for other downstream scenarios.

% Lack of the experiments studies on more sentence-level task, such as sentence similarity tasks.

% Lack of the experiments for larger size of model.

\bibliographystyle{acl_natbib}
\bibliography{main}

\end{document}